\documentclass[runningheads]{llncs}
\usepackage[T1]{fontenc}

\usepackage{graphicx}
\usepackage{xcolor}
\usepackage{textcomp}

\usepackage{amsmath,amssymb,amsfonts}
\usepackage{mathtools}

\usepackage{cite}
\usepackage{algorithm}
\usepackage{algorithmic}

\usepackage{microtype}
\usepackage{booktabs}
\usepackage{multirow}
\usepackage{subfigure}   
\usepackage{marvosym}

\usepackage[hidelinks]{hyperref}
\usepackage[capitalize,noabbrev]{cleveref}

\begin{document}
\title{Adversarial-Robustness-Guided Graph Pruning}
%
%
\author{Yongyu Wang}
%
%
\institute{}  

\maketitle              
\renewcommand{\thefootnote}{}

\renewcommand{\thefootnote}{}
\footnotetext{Correspondence to: yongyuw@mtu.edu}

\begin{abstract}
Graph learning plays a central role in many data mining and machine learning tasks, such as manifold learning, data representation and analysis, dimensionality reduction, clustering, and visualization. In this work, we propose a highly scalable, adversarial-robustness-guided graph pruning framework for learning graph topologies from data. By performing a spectral adversarial robustness evaluation, our method aims to learn sparse, undirected graphs that help the underlying algorithms resist noise and adversarial perturbations. In particular, we explicitly identify and prune edges that are most vulnerable to adversarial attacks. We use spectral clustering, one of the most representative graph-based machine learning algorithms, to evaluate the proposed framework. Compared with prior state-of-the-art graph learning approaches, the proposed method is more scalable and significantly improves both the computational efficiency and the solution quality of spectral clustering.

\end{abstract}
\section{Introduction}

Graph learning has become a core component of many modern machine learning and data mining pipelines. A common preprocessing step in these methods is to transform high-dimensional observations into a graph: each data point is modeled as a node, and edges are assigned weights that encode pairwise similarity between points. Such graphs are then exploited to capture the intrinsic structure of the data manifold and the relationships among samples\cite{jebara2009graph,liu2018learning,maier2008influence}. Despite their widespread use, learning high-quality graphs from data remains a nontrivial challenge.

A line of recent work tackles this challenge by leveraging tools from graph signal processing (GSP) to estimate sparse graph Laplacians, often with very encouraging empirical performance\cite{dong2016learning,dong2019learning,egilmez2017graph,kalofolias2017large}. For instance, \cite{rabbat2017inferring} provides error bounds for recovering sparse graphs from smooth signals, while \cite{kalofolias2017large} uses approximate nearest-neighbor (ANN) graphs to reduce the number of optimization variables, and \cite{kumar2019structured} proposes a Laplacian learning approach based on spectral constraints on the Laplacian matrix.

However, even these state-of-the-art Laplacian-based graph learning methods face serious scalability issues. For example, the optimization procedures in\cite{dong2016learning,dong2019learning,egilmez2017graph,kalofolias2016learn} incur a per-iteration time complexity of $O(N^2)$
 for $N$ data points, and require careful parameter tuning to control sparsity, which effectively restricts their applicability to relatively small datasets (typically only up to a few thousand samples). The approach in \cite{carey2017graph} constructs graphs via Isomap-based manifold embedding\cite{tenenbaum2000global}, whose manifold learning step has complexity $O(N^3)$, making it impractical for large-scale problems. Although \cite{kalofolias2017large} exploits ANN graphs to reduce the search space, the resulting method is still computationally heavy on large datasets. The Laplacian estimation strategy with spectral constraints in \cite{kumar2019structured} also presupposes access to a reasonably good initial graph structure; otherwise, searching over candidate graph structures can itself be prohibitively expensive.

This work introduces a spectral graph pruning approach for learning sparse yet robust graphs from data by leveraging a recent spectral adversarial robustness evaluation algorithm \cite{cheng2021spade} to identify and remove a small set of edges that are highly susceptible to adversarial attacks. Compared with state-of-the-art graph learning methods, the proposed approach improves graph robustness by explicitly removing non-robust edges, leading to significant gains in solution quality on benchmark data sets.

\section{Preliminaries}

Spectral graph theory studies graph properties through the eigenvalues and eigenvectors of (typically) the graph Laplacian. Consider a weighted graph \( G = (V, E, w) \), where \( V \) is the vertex set, \( E \) is the edge set, and \( w \) assigns a positive weight to each edge. The Laplacian matrix of \( G \), which is symmetric and diagonally dominant (SDD), is defined as
\begin{equation}\label{formula_laplacian}
L(p,q) = 
\begin{cases}
-w(p,q) & \text{if } (p,q) \in E, \\
\sum\limits_{(p,t) \in E} w(p,t) & \text{if } p = q, \\
0 & \text{otherwise}.
\end{cases}
\end{equation}
For improved numerical behavior and to remove the effect of scaling, one often works with the normalized Laplacian
\[
L_{\text{norm}} = D^{-1/2} L D^{-1/2},
\]
where \( D \) is the diagonal degree matrix.

In many spectral graph applications, only a small subset of eigenvalues and eigenvectors is required~\cite{chung1997spectral}. For instance, when investigating the clustering structure of a graph, the smallest eigenvalues of the Laplacian and their associated eigenvectors are typically the most informative~\cite{von2007tutorial}. Consequently, computing the full eigendecomposition is unnecessary and, for large graphs, prohibitively expensive. The Courant–Fischer minimax theorem~\cite{golub2013matrix} provides a variational characterization that allows iterative computation of eigenvalues without explicitly forming the entire spectrum. In particular, the \(k\)-th eigenvalue of a Laplacian matrix \( L \in \mathbb{R}^{|V| \times |V|} \) can be characterized as
\[
\lambda_{k}(L) = \min_{\dim(U) = k} \max_{x \in U, \, x \neq 0} \frac{x^T L x}{x^T x}.
\]

In practice, many problems involve comparing or relating two different matrices rather than analyzing a single matrix in isolation. In this setting, the Courant–Fischer theorem can be extended to a generalized form. Let \( L_X \in \mathbb{R}^{|V| \times |V|} \) and \( L_Y \in \mathbb{R}^{|V| \times |V|} \) be two Laplacian matrices such that \( \text{null}(L_Y) \subseteq \text{null}(L_X) \). For \( 1 \leq k \leq \text{rank}(L_Y) \), the \(k\)-th eigenvalue of the matrix \( L_Y^+ L_X \) admits the following variational characterization:
\[
\lambda_k(L_Y^+ L_X) = 
\min_{\substack{\dim(U) = k,\\ U \perp \text{null}(L_Y)}} \ 
\max_{x \in U,\, x \neq 0} \frac{x^T L_X x}{x^T L_Y x},
\]
where \(L_Y^+\) denotes the Moore–Penrose pseudoinverse of \(L_Y\).

\section{Method}

\begin{figure*}[htbp] 
\centering
\includegraphics[scale=0.24]{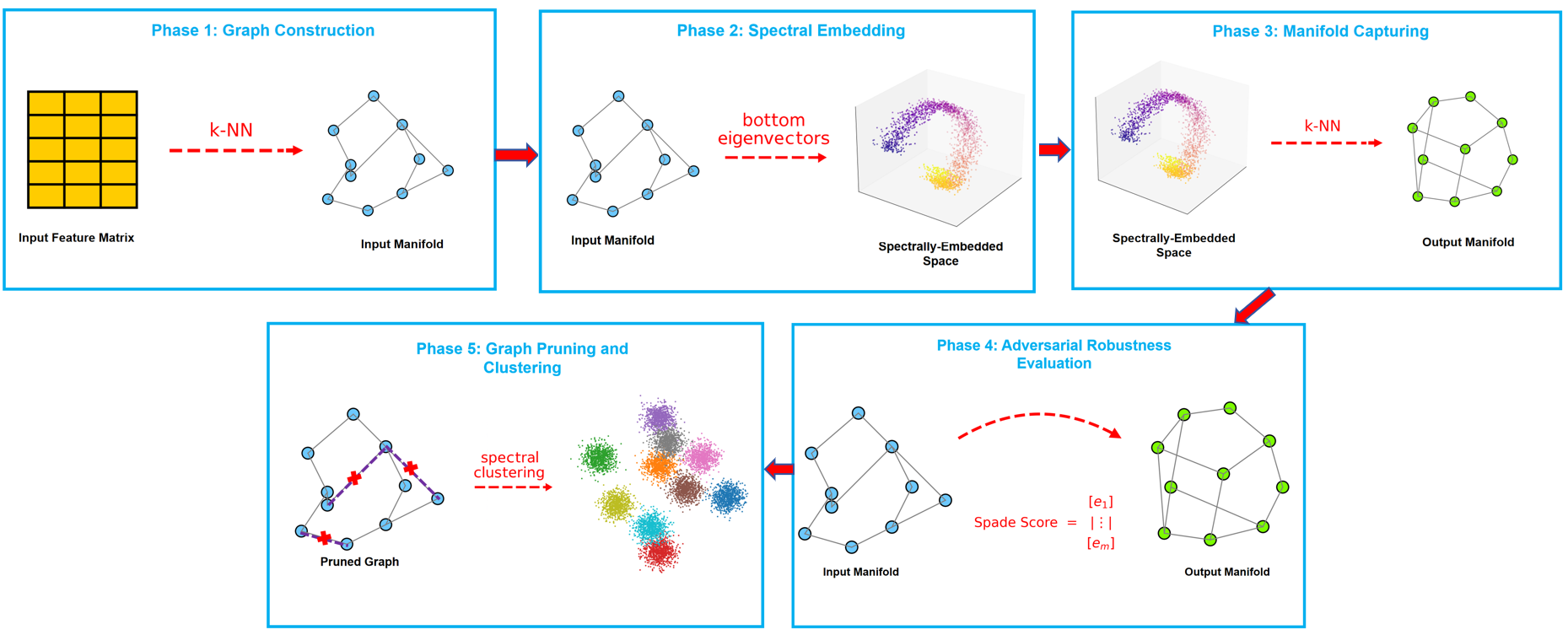}
\caption{Overview of the proposed method.}
\label{fig:flowSC}
\end{figure*}

The core idea of our method is to distinguish between edges that are robust to perturbations (robust edges) and those that are highly sensitive to noise (non-robust edges). We explicitly remove the non-robust edges from the graph and perform spectral clustering only on the graph induced by the robust edges. Specifically, the proposed method consists of the following four phases.

\subsection{Phase 1: Initial Graph Construction}

For many graph-based algorithms, the first step is to construct a graph from the given dataset~\cite{von2007tutorial}. In such a graph, each node represents a data point, and edges encode the relationships between pairs of points. In the context of spectral clustering, $k$-nearest neighbor ($k$-NN) graphs are the standard choice for this construction. In a $k$-NN graph, each node is connected to its $k$ closest neighbors, which enables the graph to capture local manifold structure effectively~\cite{roweis2000nonlinear}. This property has made $k$-NN graphs the dominant graph-construction strategy in many graph-based methods. 

However, $k$-NN graphs also tend to contain noisy or unreliable edges~\cite{yang2012affinity}. Real-world datasets often exhibit highly nonuniform and heterogeneous point distributions, so enforcing the same $k$ value for all points can be inaccurate~\cite{premachandran2013consensus}. In addition, the choice of distance metric (e.g., Euclidean distance, cosine similarity, or others) introduces further limitations~\cite{sarwar2001item}, as none of these metrics can perfectly capture the true similarity in every scenario. As a result, the neighbors identified as the $k$ closest under a given metric are not always the ones that should be connected in the underlying graph.

In this paper, we first construct a standard $k$-NN graph as the initial graph, and then refine it in subsequent stages to obtain the final graph.

\subsection{Phase 2: Spectral Graph Embedding}
To enable spectral analysis, we perform a spectral embedding of the initial $k$-NN graph by mapping it into the feature space spanned by the $K$ eigenvectors of its Laplacian matrix associated with the smallest eigenvalues. Then, we construct a k-NN graph to capture the underlying manifold of the spectrally embedded space.

\subsection{Phase 3: Adversarial Robustness Evaluation}
Let \(L_X\) and \(L_Y\) denote the Laplacian matrices of the original $k$-NN graph and the $k$-NN graph constructed in the spectrally embedded space, respectively. Following the framework in~\cite{cheng2021spade}, we use the dominant generalized eigenvalues of \(L_Y^+ L_X\) and their associated eigenvectors to assess how sensitive each sample is to noise under a given model. Specifically, we form a weighted eigensubspace matrix \(V_s \in \mathbb{R}^{|V| \times s}\) that provides a spectral embedding of the input manifold \(G_X = (V, E_X)\), where \(|V|\) denotes the number of nodes. The matrix \(V_s\) is defined as
\[
V_s = 
\begin{bmatrix}
v_1 \sqrt{\zeta_1}, & v_2 \sqrt{\zeta_2}, & \ldots, & v_s \sqrt{\zeta_s}
\end{bmatrix},
\]
where \(\zeta_1 \geq \zeta_2 \geq \cdots \geq \zeta_s\) are the largest \(s\) eigenvalues of \(L_Y^+ L_X\), and \(v_1, v_2, \ldots, v_s\) are the corresponding eigenvectors.

Each node \(p\) in \(G_X\) is then represented by the \(p\)-th row of \(V_s\), yielding an \(s\)-dimensional embedding. As described in~\cite{cheng2021spade}, the stability of an edge \((p, q) \in E_X\) is quantified by the spectral embedding distance between nodes \(p\) and \(q\):
\[
\| V_s^\top e_{p,q} \|_2^2.
\]

Based on this notion, we define a \textit{Spade score} for an edge \((p, q)\) to quantitatively measure its robustness:
\[
\text{Spade}(p, q) = \| V_s^\top e_{p,q} \|_2^2.
\]
Edges with larger \(\text{Spade}(p, q)\) values are regarded as less robust and more susceptible to perturbations along the directions determined by their incident nodes.

\subsection{Phase 4: Graph Pruning}

For the edges in the initial $k$-NN graph, we first sort them in descending order according to their Spade scores and select the top fraction to form the non-robust edge set. These edges are highly vulnerable to noise in both the graph and the feature space. We then remove them from the initial $k$-NN graph and use the resulting pruned graph as the final graph for spectral clustering.

The complete algorithm of the robust spectral clustering has been shown in Algorithm \ref{alg:flow}.

\begin{algorithm}[!h]
\caption{Robust Spectral Clustering}
\label{alg:flow}
\textbf{Input:} A data set $D$ with $N$ samples $x_1,...,x_N \in {R}^{d}$, number of clusters $k$.\\
\textbf{Output:} Clusters $C_1$,...,$C_k$.\\

\begin{algorithmic}[1] 
    \STATE Construct a $k$-nearest neighbor graph $G$ from the input data ; \\
    \STATE Compute the adjacency matrix $A_G$, and diagonal matrix $D_G$ of graph $G$; \\
    \STATE Obtain the unnormalized Laplacian matrix $L_G$=$D_G$-$A_G$;\\
    \STATE Compute the eigenvectors $u_1$,...$u_k$ that correspond to the bottom k nonzero eigenvalues of $L_G$;\\
    \STATE Construct $U \in \mathbb{R}^{n \times k}$, with $k$ eigenvectors of $L_G$ stored as columns;\\
    
    \STATE Construct a $k$-nearest neighbor graph $P = (V, E)$ from $U$.
    \STATE Compute the Laplacian matrix $\mathbf{L_P}$ for the graph $U$.
    \STATE Compute the SPADE score of each edge in graph $G$.
    \STATE Form a non-robust edge set from the edges with large SPADE scores, and remove these edges from the graph $G$ to obtain a sparse graph $T$.
    \STATE Apply standard spectral clustering on $T$ and return the result.
        
\end{algorithmic}
\end{algorithm}

\section{Experiments}
\label{sec:experiments}

\begin{table}[!htbp]
\centering
\caption{Clustering Accuracy (\%)}\label{table:compare3}
\scalebox{1}{ 
\setlength{\tabcolsep}{12pt} 
\begin{tabular}{ c c c c c c c }
\hline
Data Set & $k$-NN & Consensus & Spectral Spar &Robust Nodes&Ours \\
\hline
COIL20&75.72&81.60&76.27&67.15&\textbf{84.44}\\
PenDigits&74.36 &71.08 & 83.26  &75.71 &\textbf{84.99}\\
USPS & 64.31&68.54 & 70.74  &78.87 &\textbf{81.21}\\
MNIST &59.68 &61.09 &60.09   &\textbf{70.40} &68.21 \\
\hline
\end{tabular}}
\end{table}

\section{Conclusion}
In this work, we present a spectral graph pruning approach for graph topology learning from input data. We show that the graph topology learning problem can be addressed by improving the robustness of an initial k-NN graph, achieved by identifying and removing a small subset of edges that are most vulnerable to adversarial attacks.When compared with state-of-the-art graph learning approaches, our method exhibits better runtime scalability and leads to substantially improved solution quality in spectral clustering.
\label{sec:conclusion}


%
%
%
%

\end{document}